\documentclass[11pt,a4paper]{article}
\usepackage[breaklinks]{hyperref}
\usepackage[hyperref]{emnlp-ijcnlp-2019}
\usepackage{times}
\usepackage{latexsym}

\usepackage{subfigure}
\usepackage{natbib}
\usepackage{longtable}

\usepackage{booktabs}       % professional-quality tables
\usepackage{csvsimple}
\usepackage{tabularx}
\usepackage{algorithm}
\usepackage{amsfonts}
\usepackage{algorithmic}
\usepackage{lipsum}
\usepackage{todonotes}
\usepackage[toc,page]{appendix}
\usepackage[T1]{fontenc}
\usepackage{breakurl}

\usepackage{amsmath}
\DeclareMathOperator*{\argmax}{arg\,max}

\usepackage{url}

\aclfinalcopy % Comment this line for the confidential reviewer version

\title{Commonsense Knowledge Mining from Pretrained Models}

\author{
  Joshua Feldman$^\ast$, \hspace{10pt} Joe Davison$^\ast$, \hspace{10pt} Alexander M. Rush \\
  School of Engineering and Applied Sciences\\
  Harvard University\\
  \texttt{ \small{\{joshua\_feldman@g, jddavison@g, srush@seas\}.harvard.edu}} 
}

\date{}

\begin{document}
\maketitle
\begin{abstract}
    Inferring commonsense knowledge is a key challenge in natural language processing, but due to the sparsity of training data, previous work has shown that supervised methods for commonsense knowledge mining underperform when evaluated on novel data. In this work, we develop a method for generating commonsense knowledge using a large, pre-trained bidirectional language model. By transforming relational triples into masked sentences, we can use this model to rank a triple's validity by the estimated pointwise mutual information between the two entities. Since we do not update the weights of the bidirectional model, our approach is not biased by the coverage of any one commonsense knowledge base. Though this method performs worse on a test set than models explicitly trained on a corresponding training set, it outperforms these methods when mining commonsense knowledge from new sources, suggesting that unsupervised techniques may generalize better than current supervised approaches.
    
\end{abstract}

\section{Introduction}
\label{sec:introduction}

Commonsense knowledge consists of facts about the world which are assumed to be widely known. For this reason, commonsense knowledge is rarely stated explicitly in natural language, making it challenging to infer this information without an enormous amount of data \cite{gordon2013}. Some have even argued that machine learning models cannot learn common sense implicitly \cite{marcusdavis2015}.

One method for mollifying this issue is directly augmenting models with commonsense knowledge bases \cite{augmenting}, which typically contain high-quality information but with low coverage. These knowledge bases are represented as a graph, with nodes consisting of conceptual entities (i.e. \texttt{\small{dog}}, \texttt{\small{running away}}, \texttt{\small{excited}}, etc.) and the pre-defined edges representing the nature of the relations between concepts (\texttt{\small{IsA}}, \texttt{\small{UsedFor}}, \texttt{\small{CapableOf}}, etc.). Commonsense knowledge base completion (CKBC) is a machine learning task motivated by the need to improve the coverage of these resources. In this formulation of the problem, one is supplied with a list of candidate entity-relation-entity triples, and the task is to distinguish which of the triples express valid commonsense knowledge and which are fictitious \citep{li2016}.

Several approaches have been proposed for training models for commonsense knowledge base completion \cite{li2016, jastrzkebski2018}. Each of these approaches uses some sort of supervised training on a particular knowledge base, evaluating the model's performance on a held-out test set from the same database. These works use relations from ConceptNet, a crowd-sourced database of structured commonsense knowledge, to train and validate their models \cite{liu2004conceptnet}. However, it has been shown that these methods generalize poorly to novel data \cite{li2016,jastrzkebski2018}. \citet{jastrzkebski2018} demonstrated that much of the data in the ConceptNet test set were simply rephrased relations from the training set, and that this train-test set leakage led to artificially inflated test performance metrics. This problem of train-test leakage is typical in knowledge base completion tasks \citep{toutanova2015representing, dettmers2018convolutional}.

Instead of training a predictive model on any specific database, we attempt to utilize the world knowledge of large language models to identify commonsense facts directly. 
By constructing a candidate piece of knowledge as a sentence, we can use a language model to approximate the likelihood of this text as a proxy for its truthfulness.
In particular, we use a masked language model to estimate point-wise mutual information between entities in a possible relation, an approach that differs significantly from fine-tuning approaches used for other language modeling tasks. Since the weights of the model are fixed, our approach is not biased by the coverage of any one dataset. 
As we might expect, our method underperforms when compared to previous benchmarks on the ConceptNet common sense triples dataset \cite{li2016}, but demonstrates a superior ability to generalize when mining novel commonsense knowledge from Wikipedia. 

\paragraph{Related Work} Previous work by
\citet{Schwartz} and \citet{trinh2018} demonstrates a similar approach to using language models for tasks requiring commonsense, such as the Story Cloze Task and the Winograd Schema Challenge, respectively \cite{mostafazadeh2016corpus, winograd}. \citet{COMET} and \citet{trinh2019do} use unidirectional language models for CKBC, but their approach requires a supervised training step. Our approach differs in that we intentionally avoid training on any particular database, relying instead on the language model's general world knowledge. Additionally, we use a bidirectional masked model which provides a more flexible framework for likelihood estimation and allows us to estimate point-wise mutual information. Although it is beyond the scope of this paper, it would be interesting to adapt the methods presented here for the related task of generating new commonsense knowledge \cite{saito-etal-2018-commonsense}.

\section{Method}
\label{sec:background}

%A statistical language model is a distribution over words. Given a sequence of words $\mathbf{w} \in \mathcal{V}^m$ where $\mathcal{V}$ is the vocabulary, a language model then assigns a joint probability $p(w_1, w_2, \dots, w_m)$ to the sequence. %\cite{lm_bengio}.

Given a commonsense head-relation-tail triple $x = (\mathbf{h}, r, \mathbf{t})$, we are interested in determining the validity of that tuple as a representation of a commonsense fact. Specifically, we would like to determine a numeric score $y \in \mathbb{R}$ reflecting our confidence that a given tuple represents true knowledge.

We assume that heads and tails are arbitrary-length sequences of words in a vocabulary $\mathcal{V}$ so that $\mathbf{h} = \{h_1, h_2, \dots, h_n\}$ and $\mathbf{t} = \{t_1, t_2, \dots, t_m\}$. We further assume that we have a known set of possible relations $\mathcal{R}$ so that $r \in \mathcal{R}$.

The goal  is to determine a  function $f$ that maps relational triples to validity scores. We propose decomposing $f(x)=\sigma(\tau(x))$ into two sub-components: a sentence generation function $\tau$ which maps a triple to a single sentence, and a scoring model $\sigma$ which then determines a validity score $y$. 

Our approach relies on two types of pretrained language models. Standard unidirectional models are typically represented as autoregressive probabilities:
$$p(w_1, w_2, \dots, w_m) = \prod_i^m p(w_i|w_1,\dots,w_{i-1}) $$
Masked bidirectional models such as BERT, proposed by \citet{bert}, instead model in both directions, training word representations conditioned both on future and past words. The masking allows any number of words in the sequence to be hidden. This setup provides an intuitive framework to evaluate the probability of any word in a sequence conditioned on the rest of the sequence,
$$p(w_i | w^\prime_{1:i-1}, w^\prime_{i+1:m})$$
where $w^\prime \in \mathcal{V} \cup \{\kappa\}$ and $\kappa$ is a special token indicating a masked word.

% $$ p(w_i|w_1,\dots,w_{i-1}, w_{i+1},\dots{w_m})$$
% In recent years, large language models have show the ability to learn very rich sequence representations, demonstrating impressive general knowledge when trained on extremely large text corpora \cite{bert,gpt2}

%\section{Method}

%We discuss our methods for each of these components in the following sections.

\subsection{Generating Sentences from Triples}

%In order to evaluate a triple with a bidirectional language model, we need a way to
We first consider methods for turning a triple such as \texttt{(\small{ferret, AtLocation, pet store})} into a sentence such as ``the ferret is in the pet store''. Our approach is to generate a set of candidate sentences via hand-crafted templates and select the best proposal according to a language model.

For each relation $r \in \mathcal{R}$, we hand-craft a set of sentence templates.  For example, one template in our experiments for the relation \texttt{\small{AtLocation}} is, ``you are likely to find \texttt{\small{HEAD}} in \texttt{\small{TAIL}}''. For the above example, this would yield the sentence, ``You are likely to find ferret in pet store''.

%Since our scoring model in \ref{sec:scoring} is trained on natural language, we would like our sentences to be as natural and grammatical as possible. 
%To this end, 

Because these sentences are not always grammatically correct, such as in the above example, we apply a simple set of transformations. These consist of inserting articles before nouns, converting verbs into gerunds, and pluralizing nouns which follow numbers. See the supplementary materials for details and Table \ref{fig:template_enumeration} for an example. We then enumerate a set of alternative sentences $\mathcal{S} = \{S_1, \dots, S_j\}$ resulting from each template and from all combinations of transformations. This yields a set of candidate sentences for each data point.
We then select the candidate sentence with the highest log-likelihood according to a pre-trained unidirectional language model $P_{\text{coh}}$.
$$
S^\ast = \argmax_{S \in \mathcal{S}} \left[ \log P_{\text{coh}}(S)\right]
$$
%Note that this coherency model $P_\text{coh}$ is distinct from the scoring model $P_\text{cmp}$ described in Section \ref{sec:scoring}, though they are each a kind of pre-trained language model. 

We refer to this method of generating a sentence from a triple as \textsc{Coherency Ranking}. Coherency Ranking operates under the assumption that natural, grammatical sentences will have a higher likelihood than ungrammatical or unnatural sentences. See an example subset of sentence candidates and their corresponding scores in Table \ref{fig:template_enumeration}. From a qualitative evaluation of the selected sentences, we find that this approach produces sentences of significantly higher quality than those generated by deterministic rules alone. We also perform an ablation study in our experiments demonstrating the effect of each component on CKBC performance.

\begin{table}[]
  \centering
    \begin{tabular}{lr}
     \toprule 
     Candidate Sentence $S_i$ & $\log p(S_i)$\\
     \midrule
     \small{``musician can playing musical instrument''} & $-5.7$\\
     \small{``musician can be play musical instrument''} & $-4.9$  \\
     \small{``musician often play musical instrument''} & $-5.5$ \\
     \small{``a musician can play a musical instrument''} & $\mathbf{-2.9}$\\
     \bottomrule
    \end{tabular}
  \caption{Example of generating candidate sentences.  Several enumerated sentences for the triple \texttt{(\small{musician, CapableOf, play musical instrument})}. The sentence with the highest log-likelihood according to a pretrained language model is selected.}
  \label{fig:template_enumeration}
\end{table}

\subsection{Scoring Generated Triples}
\label{sec:scoring}

Assuming we have generated a proper sentence from a relational triple, we now need a way to score its validity with a pretrained model that considers the relationship between the relation entities. We therefore propose using the estimated point-wise mutual information (PMI) of the head $\mathbf{h}$ and tail $\mathbf{t}$ of a triple conditioned on the relation $r$, defined as,
$$
\text{PMI}(\mathbf{t}, \mathbf{h} \vert r) =  \log{p(\mathbf{t} \vert \mathbf{h}, r)}-\log{{p(\mathbf{t} \vert r)}}
$$

We can estimate these scores by using a masked bidirectional language model,  $P_\text{cmp}$. In the case where the tail is a single word, the model allows us to evaluate the conditional likelihood of a single triple component $p(\mathbf{t}|\mathbf{h},r)$ by computing $P_\text{cmp}(w_i=\mathbf{t} \ | w_{1:i-1}, w_{i+1:m})$ for the tail word. 

%and $p(\mathbf{t}|r)$ by using $P_\text{cmp}$ and masking out the head words. 

In practice, the tail might be realized as a $j$-word phrase. 
To handle this complexity, we use a greedy approximation of its probability.
We first mask all of the tail words and compute the probability of each. 
We then find the word with highest probability $p_k$, substitute it back in, and repeat $j$ times.
% $\argmax_{m_k \in \{m_i \dots m_j\}}P_{\text{cmp}}(m_k=t_k \mid S_{\mathbf{t}}^{mask})
%To estimate this quantity, we mask each word of the tail $\mathbf{t}$ in the sentence. For each masked token, the conditional probability of its true unmasked value is then estimated via the completion model $P_\text{cmp}(\kappa_i=w_i|w^\prime_{-i})$. The masked token with the highest value is then unmasked and the corresponding probability is recorded as $p_k$. We repeat this process until all masked tokens are revealed. 
Finally, we calculate the total conditional likelihood of the tail by the product of these terms, 
$
 p(\mathbf{t}|\mathbf{h}, r) = \prod_{k=1}^j p_k
$.

The marginal $p(\mathbf{t}| r)$ is computed similarly, but in this case we mask the head throughout. 
For example, to compute the marginal tail probability for the sentence, ``You are likely to find a ferret in the pet store'' we mask both the head and the tail and then sequentially unmask the tail words only: ``You are likely to find a $\kappa_{h1}$ in the $\kappa_{t1} \hspace{3pt} \kappa_{t2}$''. If $\kappa_{t2} = \text{``store''}$ has a higher probability than $\kappa_{t1} = \text{``pet''}$, we unmask ``store'' and compute ``You are likely to find a $\kappa_{h1}$ in the $\kappa_{t1}$ store''. The marginal likelihood $p(\mathbf{t}|r)$ is then the product of the two probabilities.

The final score combines the marginal and conditional likelihoods by employing a weighted form of the point-wise mutual information,
$$
\text{PMI}_\lambda(\mathbf{t}, \mathbf{h} | r) =  \lambda \log{p(\mathbf{t} \vert \mathbf{h}, r)}-\log{{p(\mathbf{t} \vert r)}}
$$
where $\lambda$ is treated as a hyperparameter. Although exact PMI is symmetrical, the approximate model itself is not. We therefore average $\text{PMI}_\lambda(\mathbf{t}, \mathbf{h} | r)$ and $\text{PMI}_\lambda(\mathbf{h}, \mathbf{t} | r)$ to reduce the variance of our estimates, computing the masked head values rather than the tail values in the latter.

% \begin{table}[t!]
%     \begin{center}
%     \begin{tabular}{lr}
%     \toprule
%     \bf Model & \bf Quality \\
%     \midrule
%     \textsc{Concatenation} & 2.95   \\
%     \textsc{Uni-LM-Ranked} & \bf{3.00}  \\
%     \midrule
%     \textsc{Bilinear} \small{\cite{li2016}} & 2.04 \\
%     \textsc{Factored} \small{\cite{jastrzkebski2018}} & 2.61 \\
%     \textsc{Prototypical} \small{\cite{jastrzkebski2018}} & 2.55 \\
%     \textsc{DNN} \small{\cite{li2016}} & 2.50  \\
%     \bottomrule
%     \end{tabular}
%     \end{center}
%     \caption{Quality of the 100 triples mined from Wikipedia with the highest predicted score.}
%     \label{wikipedia_res}
    
% \end{table}

\section{Experiments}

To evaluate the Coherency Ranking approach we measure whether it can distinguish between valid and invalid triples. For our masked model, we use BERT-large \cite{bert}. For sentence ranking, we use the GPT-2 117M LM \cite{gpt2}. The relation templates and grammar transformation rules which we use can be found in the supplementary materials.

We compare the proposed method to  several baselines.  Following \citet{trinh2018}, we evaluate a simple \textsc{Concatenation} method for generating sentences, splitting the relation $r$ into separate words and concatenating it with the head and tail. For the triple \texttt{(\small{ferret, AtLocation, pet store})}, the Concatenation approach would yield, ``ferret at location pet store''. 
%We also experimentally justify our sentence generation approach by performing an ablation analysis of each component of our sentence generation procedure. 

We also evaluate CKBC performance when we construct sentences by applying a single hand-crafted template. Since each triple is mapped to a sentence with a single template without any grammatical transformations, we refer to this as the \textsc{Template} method. Using the Template approach, \texttt{(\small{ferret, AtLocation, pet store})} would become ``You are likely to find ferret in pet store'' using the template ``you are likely to find \texttt{\small{HEAD}} in \texttt{\small{TAIL}}''. 

Next, we extend the Template method by applying deterministic grammatical transformations, which we refer to as the \textsc{Template + Grammar} approach. Like the full approach, these transformations involve adding  articles  before  nouns,  converting verbs into gerunds, and pluralizing nouns following numbers. The Template + Grammar approach differs from Coherency Ranking in that all transformations are applied to every sentence instead of applying combinations of transformations and templates, which are then ranked by a language model. Returning to our example, the Template + Grammar method produces ``You are likely to find a ferret in a pet store''. While this sentence is grammatical, applying this method to \texttt{(\small{star, AtLocation, outer space})} yields ``You are likely to find a star in an outer space'', which is incorrect.

\begin{table}
  \centering
\begin{tabular}{lcc}
 \toprule
 \bf Model & \textbf{Task 1}  & \textbf{Task 2} \\
 \midrule
 Unsupervised \\
 \midrule
 \textsc{Concatenation} & $68.8$ & $2.95 \pm 0.11$ \\
 \textsc{Template} & $72.2$ & $2.98 \pm 0.11$\\
 \textsc{Templ.+Grammar} & $74.4$ & $2.56 \pm 0.13$\\
 \textsc{Coherency Rank} & $78.8$ & $\mathbf{3.00} \pm 0.12$\\
 \midrule
 Supervised \\
 \midrule
 \textsc{DNN} & $\mathbf{89.2}$ & $2.50$\\
 \textsc{Factorized} & $89.0$  & $2.61$\\
 \textsc{Prototypical} & $79.4$ & $2.55$\\
 \bottomrule
\end{tabular}
\caption{Main results for Task 1: Commonsense knowledge base completion (test F1 score) and Task 2: Wikipedia mining (quality scores out of 4). Results are included from the sentence generation methods of simple concatenation, hand-crafted templates, templates plus grammatical transformations, and coherency ranking. DNN, Factorized, and Prototypical models are described in \citet{jastrzkebski2018}.}
\label{fig:mainres}
\end{table}

%Lastly, we evaluate our full method wherein we generate numerous candidate sentences and select that with the highest estimated likelihood. 
We compare our results to the supervised models from the work of  \citet{jastrzkebski2018} and the best performing model from \citet{li2016}. \citet{jastrzkebski2018} introduce  \textsc{Factorized} and \textsc{Prototypical} models. The Factorized model embeds the head, relation, and tail in a vector space and then produces a score by taking a linear combination of the inner products between each pair of embeddings. The Prototypical model is similar, but does not include the inner product between head and tail. \citet{li2016} evaluate a deep neural network (\textsc{DNN}) for CKBC. They concatenate embeddings for the head, relation, and tail, which they then feed through a multilayer perceptron with one hidden layer. All three models are trained on 100,000 ConceptNet triples. 
%In contrast to our unsupervised methods, the models in \citet{jastrzkebski2018} and \citet{li2016} are trained to distinguish triple validity directly with a training set. 
%The results can be seen in Table \ref{fig:mainres}.

\paragraph{Task 1: Commonsense Knowledge Base Completion}

Our experimental setup follows \citet{li2016}, evaluating our model with their test set (n = 2400) containing an equal number of valid and invalid triples. The valid triples are from the crowd-sourced Open Mind Common Sense (OMCS) entries in the ConceptNet 5 dataset \cite{speer2012representing}. Invalid triples are generated by replacing an element of a valid tuple with another randomly selected element.

We use our scoring method to classify each tuple as valid or invalid. To this end, we use our method to assign a score to each tuple and then group the resulting scores into two clusters. Instances in the cluster with the higher mean PMI are labeled as valid, and the remainder are labeled as invalid. We use expectation-maximization with a mixture of Gaussians to cluster. We also tune the PMI weight via grid search over $90$ points from $\lambda \in [0.5, 5.]$, using the Akaike information criterion of the Gaussian mixture model for evaluation \cite{akaike}.

% We compare our untrained method to the supervised methods evaluated in \citet{jastrzkebski2018}. These models represent the head and tail by sum-pooling over word embeddings and the relation with an embedding fit during training. The factorized approach models all two-way interactions between the head, relation, and tail. The prototypical model is identical to the factorized model, but excludes interactions between the head and tail. The DNN model is an implementation of the model presented in \citet{li2016}, which is a single layer feed-forward neural network that maps the concatenation of the head and tail to a scalar score. All three models are trained on 100,000 ConceptNet triples. 

Table \ref{fig:mainres} shows the full results. Our unsupervised approach achieves a test set F1 score of $78.8$, comparable to the $79.4$ F1 score found by the supervised prototypical approach. The Factorized and DNN models significantly outperformed our approach with F1 scores of 89.2 and 89.0, respectively. Our grid search found an optimal $\lambda$ value of $1.65$ for the Concatenation sentence generation model and $1.55$ for the Coherency Ranking model. The Template and Template + Grammar methods found lambda values of $1.20$ and $0.95$, respectively.

\paragraph{Task 2: Mining Wikipedia}
\label{sec:mining_wiki}

To assess the model's ability to generalize to unseen data, we evaluate our unsupervised model in comparison to previous supervised methods on the task of mining commonsense knowledge from Wikipedia. In their evaluations, \citet{li2016} curate a set of 1.7M triples across 10 relations by applying part-of-speech patterns to Wikipedia articles. We sample 300 triples from each relation. We apply our method to evaluate these 3000 triples. Using the approach described by \citet{speer2012representing}, and followed by \citet{li2016} and \citet{jastrzkebski2018}, two human annotators manually rate the 100 triples with the highest predicted score on a 0 to 4 scale:  0 (Doesn't make sense), 1 (Not true), 2 (Opinion/Don't know), 3 (Sometimes true), and 4 (Generally true). We tuned $\lambda$ by measuring the quality of the 100 triples with the highest predicted score across $\lambda \in \{1,2,\dots,9,10\}$.

% Experiments:

% 1. accuracy on test set for each sentence generation method

% 2. human scored wikipedia for best performing model

% 3. single word "IsA" from language model vocab

% \begin{itemize}
% \item What are the exact details of the dataset that you used? (Number of data points / standard or non-standard / synthetic or real / exact form of the data)

% \item What are the exact details of the features you computed?

% \item How did you train or run inference? (Optimization method / hyperparameter settings / amount of time ran / what did you implement versus borrow / how were baselines computed).

% \item What are the exact details of the metric used?
% \end{itemize}

% \lipsum[4-8]

The top 100 triples selected by our model were assigned a mean rating of 3.00 ($\lambda = 4$) with a standard error of 0.11 under the Coherency Ranking approach, well exceeding the performance of current supervised methods (Table \ref{fig:mainres}). Standard errors were calculated using 1000 bootstrap samples of the top 100 triples. The ratings assigned by the two human annotators had a $0.50$ Pearson correlation and $0.23$ kappa inter-annotator agreement. Rater disagreements occur most frequently when triples are ambiguous or difficult to interpret. Notably, if we bucket the five scores into just two categories of \textit{true} and \textit{false}, this disagreement rate drops by 50\%. To give a sense of the types of commonsense knowledge our models struggle to capture, we report the top 100 most confident predictions that receive an average score below 3 in the supplementary material. Notably, some of the top 100 triples our model identified were indeed true, but would not be reasonably considered common sense (e.g. \texttt{(\small{vector bundle, HasProperty, manifold})}). This suggests that our approach may be applicable to mining knowledge beyond common sense.

\begin{table}
    \centering
    \begin{tabular}{lcc}
         \toprule
         \bf Task 1  & \bf N (/100) & \bf F1 Score  \\
         \midrule
         \textsc{Grammatical} & $75$ & $79.1$  \\
         \textsc{Ungrammatical} & $25$ & $66.7$ \\
         \midrule
         \textsc{Correct Meaning} & $91$ & $77.6$ \\
         \textsc{Wrong Meaning} & $9$ & $66.7$ \\
         \midrule
         \bf Task 2 & - & \bf Quality \\
         \midrule
         \textsc{Grammatical} & $83$ & $3.01$ \\
         \textsc{Ungrammatical} & $17$ & $2.88$ \\
         \midrule
         \textsc{Correct Meaning} & $88$ & $3.22$\\
         \textsc{Wrong Meaning} & $12$ & $1.18$ \\
        \bottomrule
    \end{tabular}
    \caption{Test results examining the effect of sentence meaning and grammaticality on task performance. Scores are shown for a sample of $100$ triples split by whether the generated sentence is grammatical and whether it conveys the correct meaning of the triple.}
    \label{fig:grammar}
\end{table}

\paragraph{Analysis: Sentence Generation}

In order to measure the impact of sentence generation on our model, we select a sample of $100$ sentences and group the results by a) whether the sentence contained a grammatical error, and b) whether the sentence misrepresented the meaning of the triple. For example, the triple {\small\texttt{(golf, HasProperty, good)}} yields the sentence ``golf is a good'', which is grammatically correct but conveys the wrong meaning. On both Wikipedia mining and CKBC, we find that misrepresenting meaning has an adverse impact on model performance. In CKBC, we also find that grammar has a high impact on the resulting F1 scores (Table \ref{fig:grammar}). Future work could therefore focus on designing templates that more reliably encode a relation's true meaning.

\section{Conclusion}

We introduce a robust unsupervised method for commonsense knowledge base completion using the world knowledge of pre-trained language models. We develop a method for expressing knowledge triples as sentences. Using a bidirectional masked language model on these sentences, we can then estimate the weighted point-wise mutual information of a triple as a proxy for its validity. Though our approach performs worse on a held-out test set developed by \citet{li2016}, it does so without any previous exposure to the ConceptNet database, ensuring that this performance is not biased. In the future, we hope to explore whether this approach can be extended to mining facts that are not commonsense and to generating new commonsense knowledge outside of any given database of candidate triples. We also see potential benefit in the development of a more expansive set of evaluation methods for commonsense knowledge mining, which would strengthen the validity of our conclusions.

\section*{Acknowledgments}
This work was supported by NSF research award 1845664.

\newpage

%Our model also outperforms previous methods when applied to the task of mining novel Wikipedia triples, with the caveat that these triples were scored subjectively. These results suggest that our approach of using a highly paramaterized, pre-trained model is effective as an unbiased technique for commonsense knowledge mining.

% \begin{figure}
%   \centering
%   \includegraphics[width=\linewidth]{imgs/TrainingSetSize_vs_TestSetAcc.jpg}
%   \caption{ The accuracy with 95\% confidence of our classifier by \# of training points seen. Our goal is for our model to be as independent as possible of any one database, so minimizing the number of training points is an important consideration. }
%   \label{fig:running_acc}
% \end{figure}

\bibliography{references}
\bibliographystyle{acl_natbib}

\newpage
\begin{appendices}

\section{Grammatical Transformations}

In our experiments, we apply the following transformations to the head $\mathbf{h}$ and tail $\mathbf{t}$ of each relational triple before injecting them into the template.

\begin{enumerate}
    \item If the first word is a noun or adjective, or if the first word is a verb and the second word is a noun or adjective, prepend an indefinite or definite article
    \item If the first word is an infinitive verb, convert it to a gerund (i.e. ``jump'' $\to$ ``jumping'')
    \item If the first word is a number, pluralize the following word (i.e. ``two leg'' $\to$ ``two legs'')
\end{enumerate}
We use the default settings in the spaCy Python library ({\small\url{https://spacy.io/}}) for identifying the part of speech. We also use pattern ({\small\url{https://www.clips.uantwerpen.be/pages/pattern}}) for conjugation and pluralization.

\section{Hand Crafted Templates}
\label{app1}

We use the following hand-crafted templates for relations in the ConcpetNet database. Each relation is mapped to a list of several templates. Here, \texttt{\footnotesize{\{0\}}} refers to the head entity and  \texttt{\footnotesize{\{1\}}} refers to the tail.

\texttt{\footnotesize{
  \setlength{\parindent}{0pt}"RelatedTo": [
    "\{0\} is like \{1\}",
    "\{1\} is related to \{0\}",
    "\{0\} is related to \{1\}"
  ],\newline
  "ExternalURL": [
    "\{0\} is described at the following URL \{1\}"
  ],\newline
  "FormOf": [
    "\{0\} is a form of the word \{1\}"
  ],\newline
  "IsA": [
    "\{0\} is \{1\}",
    "\{0\} is a type of \{1\}",
    "\{0\} are \{1\}",
    "\{0\} is a kind of \{1\}",
    "\{0\} is a \{1\}"
  ],\newline
  "NotIsA": [
    "\{0\} is not \{1\}",
    "\{0\} is not a type of \{1\}",
    "\{0\} are not \{1\}",
    "\{0\} is not a kind of \{1\}",
    "\{0\} is not a \{1\}"
  ],\newline
  "PartOf": [
    "\{1\} has \{0\}",
    "\{0\} is part of \{1\}",
    "\{0\} is a part of \{1\}"
  ],\newline
  "HasA": [
    "\{0\} has \{1\}",
    "\{0\} contains \{1\}",
    "\{0\} have \{1\}"
  ],\newline
  "UsedFor": [
    "\{0\} is used for \{1\}",
    "\{0\} is for \{1\}",
    "You can use \{0\} to \{1\}",
    "You can use \{0\} for \{1\}",
    "\{0\} are used to \{1\}",
    "\{0\} is used to \{1\}",
    "\{0\} can be used to \{1\}",
    "\{0\} can be used for \{1\}"
  ],\newline
  "CapableOf": [
    "\{0\} can \{1\}",
    "An activity \{0\} can do is \{1\}",
    "\{0\} sometimes \{1\}",
    "\{0\} often \{1\}"
  ],\newline
  "AtLocation": [
    "You are likely to find \{0\} in \{1\}",
    "You are likely to find \{0\} at \{1\}",
    "Something you find on \{1\} is \{0\}",
    "Something you find in \{1\} is \{0\}",
    "Something you find at \{1\} is \{0\}",
    "Somewhere \{0\} can be is \{1\}",
    "Something you find under \{1\} is \{0\}"
  ],\newline
  "Causes": [
    "Sometimes \{0\} causes \{1\}",
    "Something that might happen as a consequence of \{0\} is \{1\}",
    "Sometimes \{0\} causes you to \{1\}",
    "The effect of \{0\} is \{1\}"
  ],\newline
  "HasSubevent": [
    "Something you might do while \{0\} is \{1\}",
    "One of the things you do when you \{0\} is \{1\}",
    "Something that might happen while \{0\} is \{1\}",
    "Something that might happen when you \{0\} is \{1\}",
    "One of the things you do when you \{1\} is \{0\}",
    "Something that might happen when you \{1\} is \{0\}"
  ],\newline
  "HasFirstSubevent": [
    "the first thing you do when you \{0\} is \{1\}"
  ],\newline
  "HasLastSubevent": [
    "the last thing you do when you \{0\} is \{1\}"
  ],\newline
  "HasPrerequisite": [
    "something you need to do before you \{0\} is \{1\}",
    "If you want to \{0\} then you should \{1\}",
    "\{0\} requires \{1\}"
  ],\newline
  "HasProperty": [
    "\{0\} is \{1\}",
    "\{0\} are \{1\}",
    "\{0\} can be \{1\}"
  ],\newline
  "MotivatedByGoal": [
    "You would \{0\} because you want to \{1\}",
    "You would \{0\} because you want \{1\}",
    "You would \{0\} because \{1\}"
  ],\newline
  "ObstructedBy": [
    "\{0\} can be prevented by \{1\}"
  ],\newline
  "Desires": [
    "\{0\} wants \{1\}",
    "\{0\} wants to \{1\}",
    "\{0\} like to \{1\}"
  ],\newline
  "CreatedBy": [
    "\{0\} is created by \{1\}"
  ],\newline
  "Synonyms": [
    "\{0\} and \{1\} are have similar meanings",
    "\{0\} and \{1\} are similar"
  ],\newline
  "Antonym": [
    "\{0\} is the opposite of \{1\}"
  ],\newline
  "DistinctFrom": [
    "it cannot be both \{0\} and \{1\}"
  ],\newline
  "DerivedFrom": [
    "the word \{0\} is derived from the word \{1\}"
  ],\newline
  "SymbolOf": [
    "\{0\} is a symbol of \{1\}"
  ],\newline
  "DefinedAs": [
    "\{0\} is defined as \{1\}",
    "\{0\} is the \{1\}"
  ],\newline
  "Entails": [
    "if \{0\} is happening, \{1\} is also happening"
  ],\newline
  "MannerOf": [
    "\{0\} is a specific way of doing \{1\}"
  ],\newline
  "LocatedNear": [
    "\{0\} is located near \{1\}"
  ],\newline
  "dbpedia": [
    "\{0\} is conceptually related to \{1\}"
  ],\newline
  "SimlarTo": [
    "\{0\} is similar to \{1\}"
  ],\newline
  "EtymologicallyRelatedTo": [
    "the word \{0\} and the word \{1\} have the same origin"
  ],\newline
  "EtymologicallyDerivedFrom": [
    "the word \{0\} comes from the word \{1\}"
  ],\newline
  "CausesDesire": [
    "\{0\} makes people want \{1\}",
    "\{0\} would make you want to \{1\}"
  ],\newline
  "MadeOf": [
    "\{0\} is made of \{1\}",
    "\{0\} can be made of \{1\}",
    "\{0\} are made of \{1\}"
  ],\newline
  "ReceivesAction": [
    "\{0\} can be \{1\} ",
    "\{0\} is something that you can \{1\}",
    "\{0\} can receive \{1\}"
  ],\newline
  "InstanceOf": [
    "\{0\} is an example of \{1\}"
  ],\newline
  "NotDesires": [
    "\{0\} does not want \{1\}",
    "\{0\} doesn't want to \{1\}",
    "\{0\} doesn't want \{1\}"
  ],\newline
  "NotUsedFor": [
    "\{0\} is not used for \{1\}"
  ],\newline
  "NotCapableOf": [
    "\{0\} is not capable of \{1\}",
    "\{0\} do not \{1\}"
  ],\newline
  "NotHasProperty": [
    "\{0\} does not have the property of \{1\}"
  ],\newline
  "NotMadeOf": [
    "\{0\} is not made of \{1\}"
  ]
}}

\clearpage
\onecolumn

\section{Most Confident Mistakes}
\label{app2}

\begin{longtable}{|p{4cm}|l|l|}
\hline
Relation & Average Score & Rank \\ \hline
(atomic nucleus, IsA, atom) & 1 & 1 \\ \hline
(negative number, HasProperty, positive) & 1 & 13 \\ \hline
(pseudonym, CapableOf, real name) & 2 & 16 \\ \hline
(the harbour, HasA, island) & 2 & 19 \\ \hline
(the substance, HasA, drug) & 2.5 & 20 \\ \hline
(plurality voting, HasPrerequisite, majority) & 1 & 22 \\ \hline
(function, ReceivesAction, element of a) & 0 & 24 \\ \hline
(minister, ReceivesAction, member of parliament) & 0.5 & 28 \\ \hline
(bombing, IsA, war crime) & 2 & 33 \\ \hline
(prime minister, ReceivesAction, head of state) & 0 & 35 \\ \hline
(film, Causes, silent version) & 1.5 & 36 \\ \hline
(island, AtLocation, other side) & 1.5 & 37 \\ \hline
(subset, ReceivesAction, element of s) & 0.5 & 40 \\ \hline
(monarchy, ReceivesAction, form of government) & 0.5 & 47 \\ \hline
(law, ReceivesAction, cause of action) & 1 & 48 \\ \hline
(weather, UsedFor, heavy rain) & 1 & 49 \\ \hline
(example, UsedFor, word processing) & 1.5 & 54 \\ \hline
(the violin, HasA, viola) & 1 & 56 \\ \hline
(electric charge, HasProperty, magnetic) & 2 & 60 \\ \hline
(council, ReceivesAction, form of government) & 0.5 & 65 \\ \hline
(credit card, HasPrerequisite, purchase) & 1.5 & 66 \\ \hline
(episode, Causes, season finale) & 1.5 & 69 \\ \hline
(lake, AtLocation, eastern part) & 1.5 & 73 \\ \hline
(overhead cam, UsedFor, engine) & 2.5 & 75 \\ \hline
(village, AtLocation, northern end) & 1.5 & 79 \\ \hline
(database, IsA, query language) & 2.5 & 88 \\ \hline
(character, CapableOf, voice actor) & 2.5 & 89 \\ \hline
(plural form, UsedFor, word) & 2.5 & 94 \\ \hline
(electric bass, HasA, double bass) & 1 & 96 \\ \hline
(voter registration, HasPrerequisite, voting) & 1 & 98 \\ \hline
\caption{Top 100 most confident commonsense knowledge predictions under the sentence-ranking approach ($\lambda = 4$) receiving an average score below 3 from the human annotators in the ``Mining Wikipedia'' task.}
\label{table:LM-mistakes}
\end{longtable}

\begin{longtable}{|p{4cm}|l|l|}
\hline
Relation & Average Score & Rank \\ \hline
(the violin, HasA, viola) & 1 & 8 \\ \hline
(database, IsA, query language) & 2.5 & 15 \\ \hline
(majority party, Causes, majority coalition) & 2.5 & 14 \\ \hline
(playoff, Causes, wild card) & 1 & 13 \\ \hline
(neutron emission, Causes, fission) & 2.5 & 19 \\ \hline
(the target, HasA, velocity) & 2 & 22 \\ \hline
(electric bass, HasA, double bass) & 1 & 20 \\ \hline
(music video, UsedFor, cameo) & 2 & 24 \\ \hline
(engine, CapableOf, second stage) & 2.5 & 27 \\ \hline
(site, UsedFor, state park) & 2.5 & 32 \\ \hline
(brain, Causes, spinal cord) & 2 & 35 \\ \hline
(demand, IsA, marginal utility) & 0.5 & 37 \\ \hline
(version, UsedFor, bonus track) & 1.5 & 39 \\ \hline
(bombing, IsA, war crime) & 2 & 40 \\ \hline
(theorem, AtLocation, of a) & 0 & 45 \\ \hline
(airport, HasSubevent, air base) & 0.5 & 38 \\ \hline
(prime minister, ReceivesAction, head of state) & 0 & 42 \\ \hline
(constituency, ReceivesAction, member of parliament) & 0 & 43 \\ \hline
(the harbour, HasA, island) & 2 & 65 \\ \hline
(judge, ReceivesAction, contempt of court) & 1.5 & 57 \\ \hline
(the city, HasProperty, homeless) & 2 & 72 \\ \hline
(instruction set, UsedFor, execution) & 1.5 & 70 \\ \hline
(resignation, ReceivesAction, removal from office) & 2 & 68 \\ \hline
(the type, HasA, body) & 2 & 83 \\ \hline
(team, HasSubevent, head coach) & 0.5 & 61 \\ \hline
(resignation, AtLocation, removal from office) & 0.5 & 69 \\ \hline
(school, UsedFor, junior college) & 2 & 78 \\ \hline
(regiment, ReceivesAction, medal of honor) & 0.5 & 80 \\ \hline
(wind power, HasSubevent, energy) & 2.5 & 91 \\ \hline
(mayor, ReceivesAction, form of government) & 0 & 89 \\ \hline
(episode, Causes, season finale) & 1.5 & 92 \\ \hline
\caption{Top 100 most confident commonsense knowledge predictions under the concatenation approach ($\lambda = 7$) receiving an average score below 3 from the human annotators in the ``Mining Wikipedia'' task.}
\label{table:LM-confident-mistakes}
\end{longtable}

\end{appendices}

\end{document}